\documentclass[runningheads]{llncs}

\usepackage{eccv}

\usepackage{tabularx}
\usepackage{eccvabbrv}

\usepackage{graphicx}
\usepackage{xcolor}
\usepackage{color}
\usepackage{color, colortbl}
\usepackage{subcaption} 
\usepackage{booktabs} 
\usepackage{array}    
\usepackage{makecell} 
\usepackage{multirow} 
\usepackage{caption}  
\usepackage{url}      
\usepackage{authblk}
\usepackage{hyperref} 

\usepackage{times}

\newcommand{\chighlight}[2]{\colorbox{#1}{$\displaystyle #2$}}

\usepackage[accsupp]{axessibility}  

\usepackage{hyperref}

\usepackage{orcidlink}

\begin{document}

\title{JoyStreamer: Unlocking Highly Expressive Avatars via Harmonized Text-Audio Conditioning}

\titlerunning{JoyStreamer}

\author{Ruikui Wang \and Jinheng Feng \and Lang Tian \and Huaishao Luo \and Chaochao Li \and Liangbo Zhou \and Huan Zhang\thanks{Project Leader} \and 
Youzheng Wu \and
Xiaodong He}
\institute{JD Technology}


\authorrunning{Ruikui Wang et al.}


\maketitle

\begin{figure}[h]
  \vspace{-0.5cm}
  \centering 
  \includegraphics[width=0.9\textwidth,trim=0.0cm 0.0cm 0.0cm 0.0cm,clip]{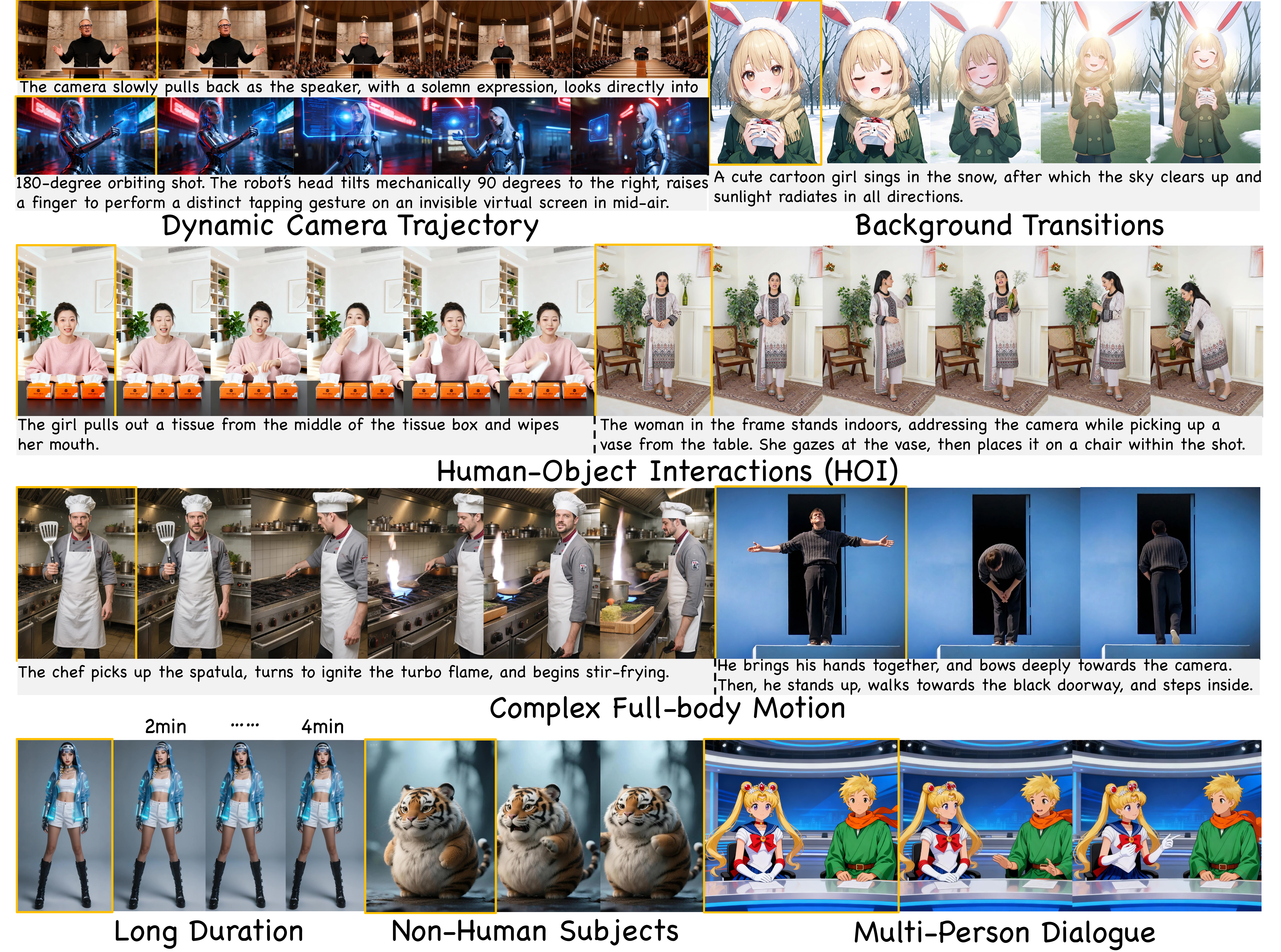}
  \vspace{-8pt}
  \caption{JoyStreamer model can generate vivid avatar videos with the input prompts involving complex motion patterns (The reference images are marked with orange borders, and the text prompt is displayed below the video).}
  \label{fig:intro-00}
\end{figure}
\begin{figure}[h]
    \vspace{-1.6cm}
    \centering
    \begin{minipage}{0.43\textwidth}
        \centering
        \includegraphics[width=\linewidth]{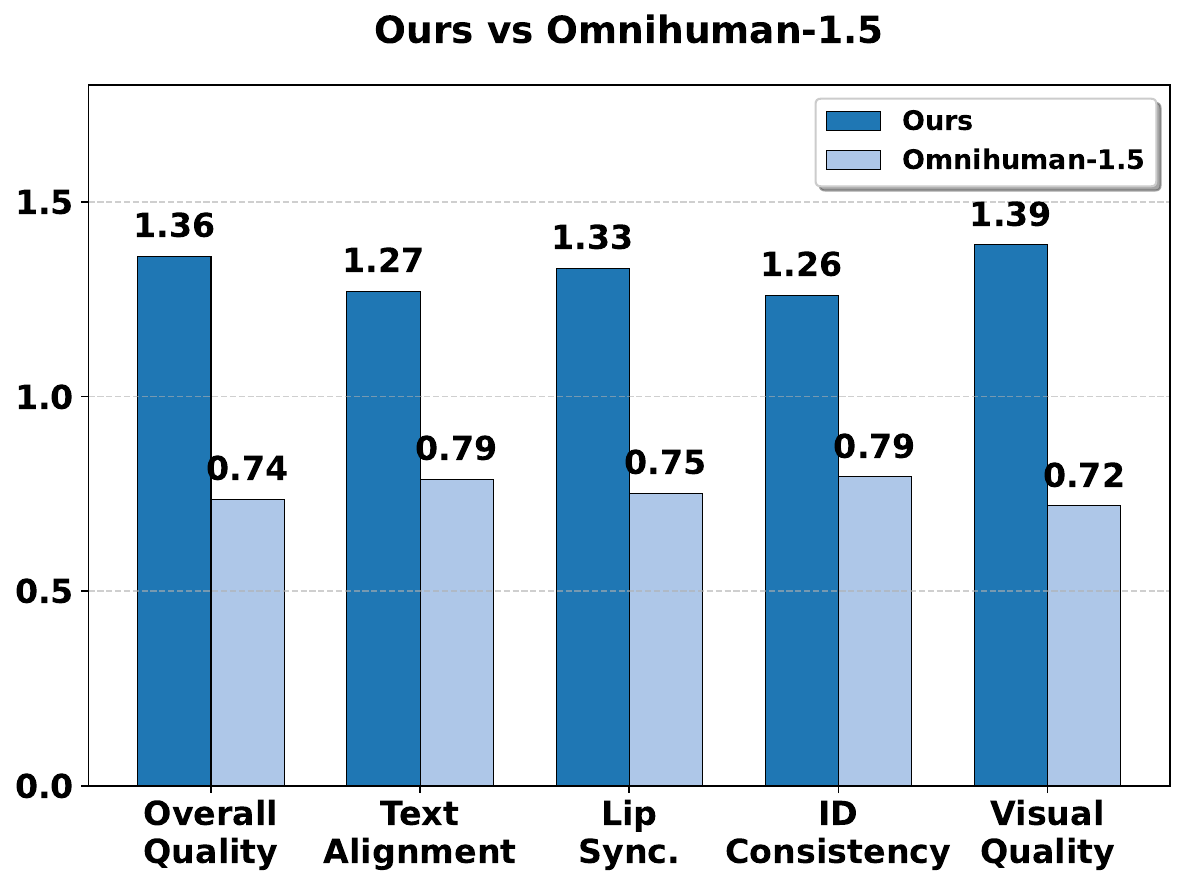}
    \end{minipage}
    \begin{minipage}{0.43\textwidth}
        \centering
        \includegraphics[width=\linewidth]{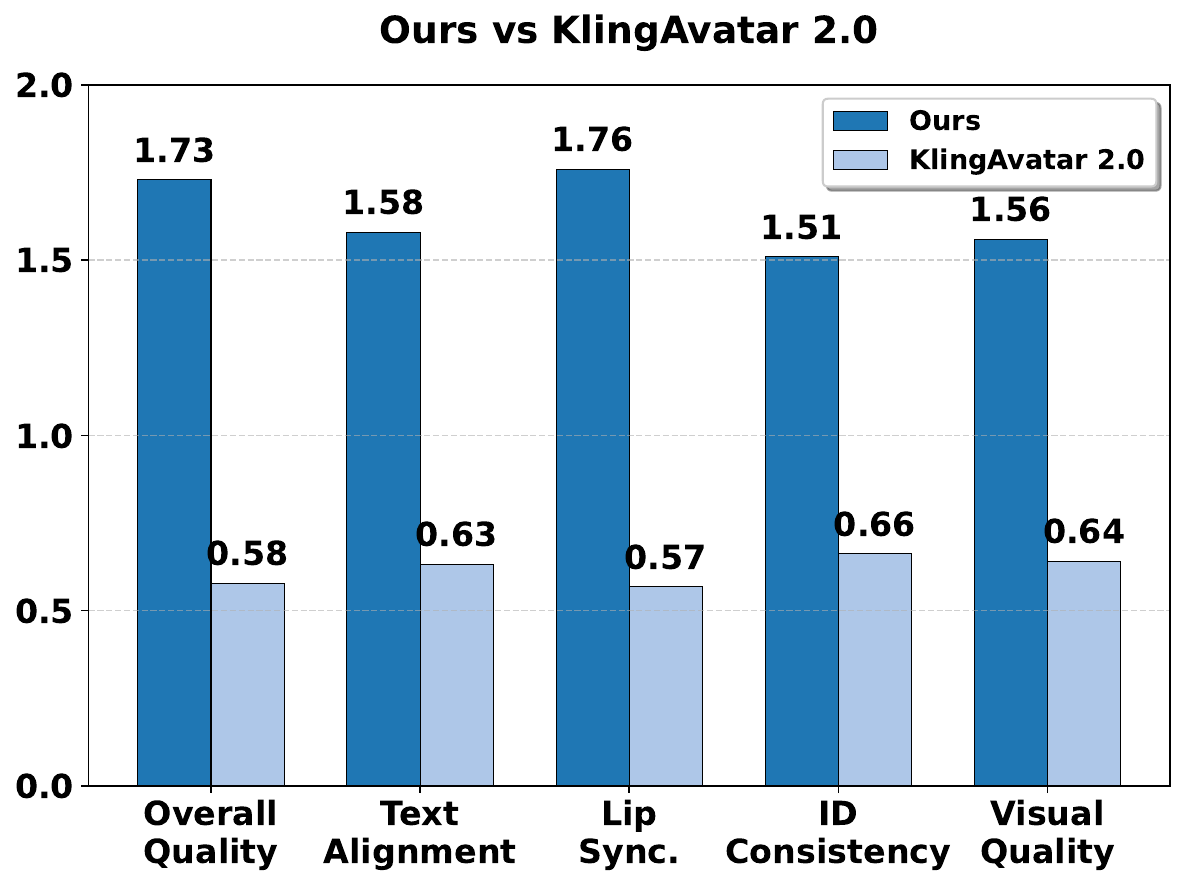}
    \end{minipage}
    \vspace{-8pt}
    \caption{Visualization of GSB results comparing our method with OmniHuman-1.5 and KlingAvatar 2.0 across various evaluation dimensions.}
    \label{fig:comparison-with-sota}
    \vspace{-1cm}
\end{figure}

\begin{abstract}
Existing video avatar models have demonstrated impressive capabilities in scenarios such as talking, public speaking, and singing. However, the majority of these methods exhibit limited alignment with respect to text instructions, particularly when the prompts involve complex elements including large full-body movement, dynamic camera trajectory, background transitions, or human-object interactions. To break out this limitation, we present JoyStreamer, a framework capable of generating long duration avatar videos, featuring two key technical innovations. Firstly, we introduce a twin-teacher enhanced training algorithm that enables the model to transfer inherent text-controllability from the foundation model while simultaneously learning audio–visual synchronization. Secondly, during training, we dynamically modulate the strength of multi-modal conditions (e.g., audio and text) based on the distinct denoising timestep, aiming to mitigate conflicts between the heterogeneous conditioning signals. These two key designs serve to substantially expand the avatar model's capacity to generate natural, temporally coherent full-body motions and dynamic camera movements as well as preserve the basic avatar capabilities, such as accurate lip-sync and identity consistency. GSB evaluation results demonstrate that our JoyStreamer model outperforms the state-of-the-art models such as Omnihuman-1.5 and KlingAvatar 2.0. Moreover, our approach enables complex applications including multi-person dialogues and non-human subjects role-playing. Some video samples are provided on \url{https://joystreamer.github.io/}.

  \keywords{Audio–Visual Synchronization \and Video Avatar Generation \and Text Alignment}
\end{abstract}

\section{Introduction}
\label{sec:intro}
\begin{figure}[tb]
    \centering 
    \includegraphics[width=0.91\textwidth,trim=0cm 0cm 0cm 0cm,clip]{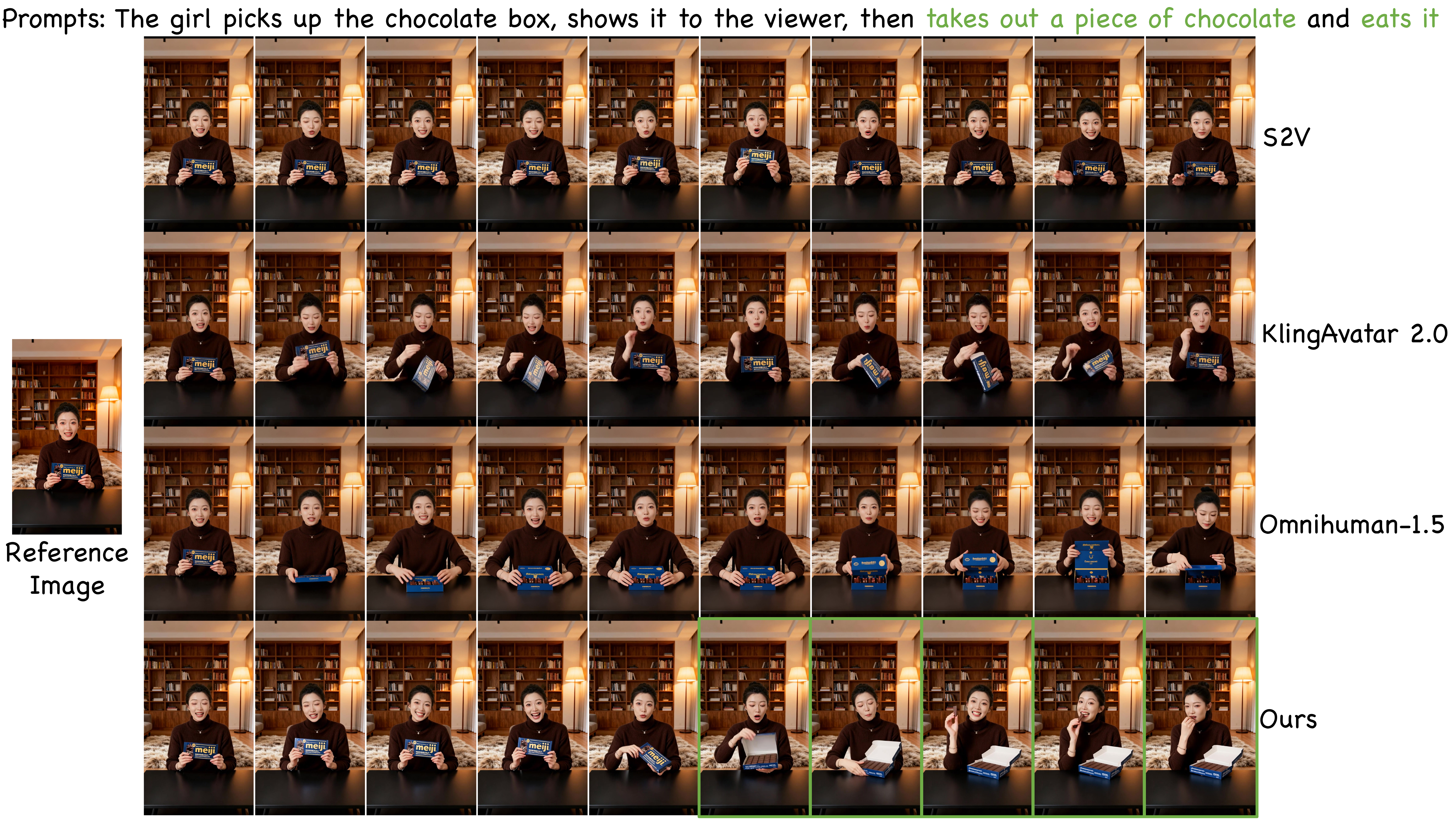} \\
    \includegraphics[width=0.91\textwidth,trim=0cm 0cm 0cm 0cm,clip]{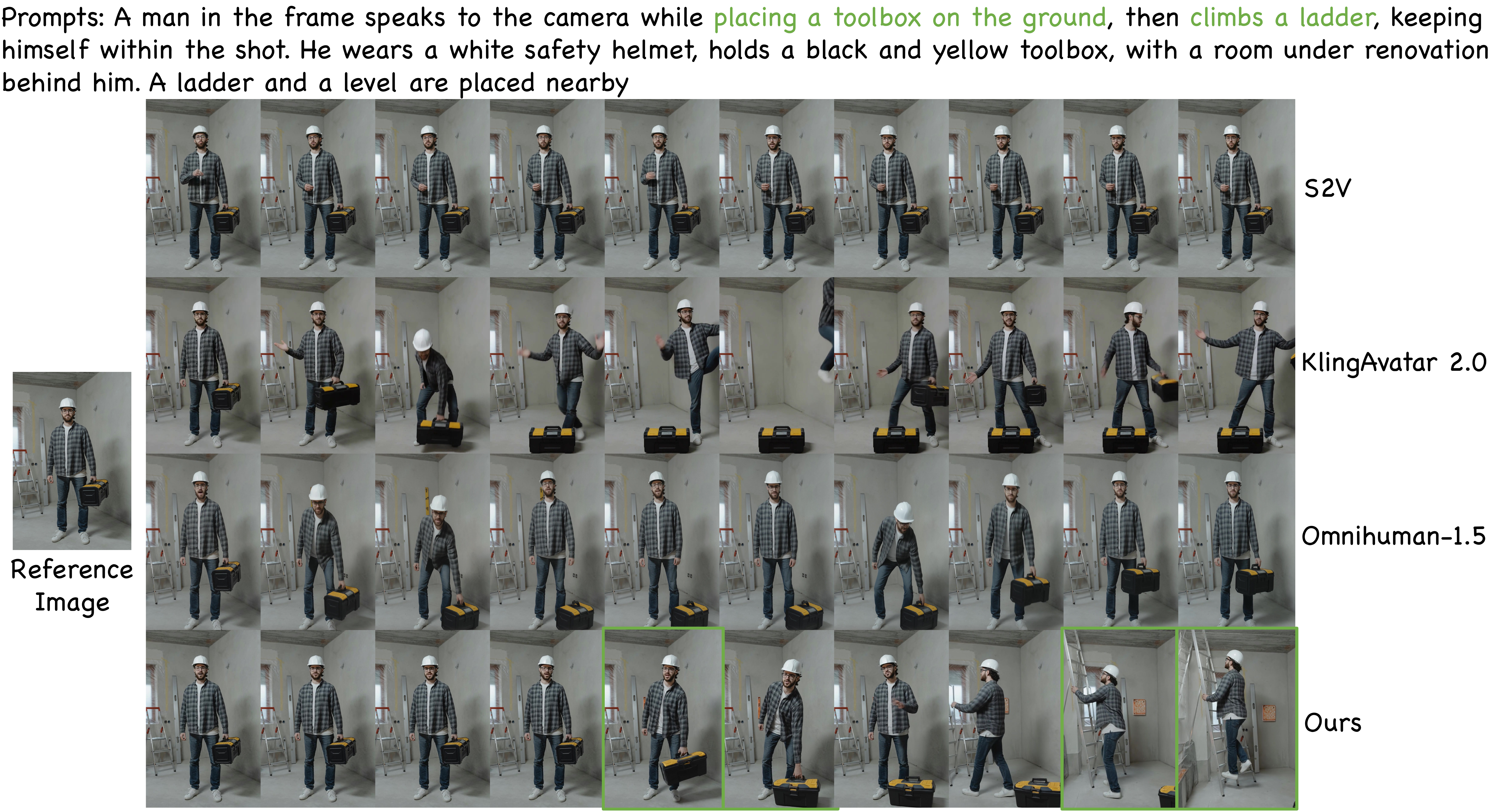}
    \caption{Some notable phenomena in existing avatar models: insufficient control over text prompts (Key actions in the prompts are highlighted in green and generated frames that faithfully adhere to the input prompt are marked with green borders).}
    \label{fig:complex-text-aligment}
    \vspace{-0.5cm}
\end{figure}

Current audio-driven avatar video generation models \cite{cui2025high,meng2025echomimicv2,tian2025emo2,kong2025let,wang2025fantasytalking,wang2025fantasytalking2,meng2025echomimicv3,yang2025infinitetalk,tu2025stableavatar,gan2025omniavatar,li2025joyavatar,lin2025omnihuman} have made significant progress in aspects such as audio-lip synchronization and identity consistency. However, they often produce motions that are limited in diversity and appear rigid. Although text prompts can be employed to control the motions of avatars, most of these approaches often suffer from ineffective prompt responsiveness—particularly when prompts involve complex motion patterns, such as specific character actions or camera movements within a scene. Most recently, several studies have begun exploring ways to enhance the text-alignment of avatar video generation models. Typically, Wan-S2V \cite{gao2025wan} proposes to train the model on carefully curated data, equipped with faithful textual captioning, to maintain strong text controllability. However, in this approach, the use of the reference latent severely restricts the dynamics of the motion. KlingAvatar \cite{ding2025kling} reformulates this problem as a first-last frame conditioning model; however, this approach tends to produce repetitive motion patterns or motions that violate physical plausibility, which hinders the model from generating motions that faithfully align with specific text prompts. Omnihuman-1.5 \cite{jiang2025omnihuman} and KlingAvatar 2.0 \cite{team2025klingavatar} exhibit superior text-instruction following performance relative to most open-source counterparts. Nevertheless, their performance driven by some complex text instructions still leaves substantial room for improvement, particularly those involving human-object interactions (See Figure~\ref{fig:complex-text-aligment}).


Rethinking the training process of avatar models, we identify two primary factors that limit the text-controllability: 1) \textbf{Data Bias}: the majority of training data originates from relatively static scenarios such as speeches, talk shows, and television hosting, which lack complex elements like dynamic camera movements or large-movement body motions. As training progresses, avatar models progressively overfit to the narrow distribution of speaking-style motion patterns, which consequently undermine their ability to faithfully follow text instructions specifying large-movement motion; 2) \textbf{Inter-Modality Conflicts}: The general avatar model is simultaneously conditioned on reference image, audio, and text driven signals. The audio signal predominantly governs rhythmic motion dynamics, and the reference image implicitly constrains the motion range. Consequently, both signals introduce competing inductive biases that interfere with the text prompts control over motion patterns.

To this end, we propose a new avatar generation framework specifically designed to resolve the aforementioned issues. For the issue of data bias, we propose to introduce a teacher model that provides persistent text-aligned supervision throughout avatar training. This strategy obviates the high cost of curating new diverse-style datasets while preserves the model’s text-controllability, particularly in motion pattern control. Considering that teacher models are typically integrated in post-training phases, e.g., commonly used DMD (Distribution Matching Distillation) \cite{yin2024one}, we lean to introduce this novel strategy at this phase. Specifically, under the DMD training framework, we employ a basic avatar model as the original teacher and introduce a pretrained video foundation model as another teacher, such as Wan2.2-I2V foundation model \cite{wan2025wan}. The original teacher provides audio-driven guidance, while the another teacher specializes in text-driven guidance. This Twin-Teacher paradigm enables student avatar model to effectively disentangle and leverage modality-specific supervision (for more details, see Sec.~\ref{sec_post-training}). For the issue of inter-modality conflicts, we additionally introduce a dynamic classifier-free guidance (CFG) strategy during the DMD training phase. Recognizing that global motion patterns are predominantly established in the high-noise (early) denoising timesteps, while fine-grained visual details are refined in the low-noise (later) timesteps, we adaptively modulate the CFG scales across modalities. Specifically, we assign a lower CFG scale to audio in early timesteps to prioritize text-driven semantic alignment, and gradually increase the audio CFG scale in later timesteps to ensure precise lip-sync fidelity. This training scheduling based on timestep allows each modality to exert its influence predominantly in complementary stages of the generation process, thereby mitigating inter-modality conflicts and enhancing overall controllability (for more details, see Sec.~\ref{sec_post-training}). As a result of the aforementioned designs, our avatar model offers a unified solution that effectively addresses the limitation of weak text-controllability.

We summarize our main contributions as follows:

1) We propose a Twin-Teacher Enhanced DMD post-training approach, without introducing any new training data, which effectively endows the avatar model with precise text-instruction following capabilities.

2) We incorporate a dynamic CFG strategy into the training process, wherein audio and text conditions are separately modulated according to the denoising timestep interval. This adaptive conditioning mechanism effectively mitigates the conflict between the distinct control modalities.

3) Our avatar model achieves state-of-the-art performance across various dimensions, including text alignment, lip synchronization, identity consistency, and visual quality, and enables diverse applications such as multi-character conversation and cinematic content creation.

\section{Related Work}

\textbf{Video Generation.}
Leveraging the powerful generative capability of diffusion models \cite{ho2020denoising}, video generation has recently witnessed significant breakthroughs. Early approaches \cite{guoanimatediff,singermake} commonly employed a 2D U-Net architecture \cite{ronneberger2015u}, wherein the model was first pretrained on large-scale image datasets and then extended with temporal modeling modules—such as motion module—before being fine-tuned on video data. Although this strategy effectively exploits strong image priors learned from static images, it frequently results in suboptimal temporal consistency across generated frames, manifesting as flickering or discontinuous motion. Subsequently, the SVD (Stable Video Diffusion) approach \cite{blattmann2023stable} integrates temporal and spatial information within a unified architecture, enabling bidirectional information exchange across both dimensions and thereby generating videos with significantly improved temporal coherence. Recently, the dominant paradigm in video generation has shifted toward DiT (Diffusion Transformer)-based architectures \cite{chen2024gentron,kong2024hunyuanvideo,menapace2024snap,polyak2024movie,yangcogvideox,zheng2024open,wan2025wan}. These methods typically partition input video data into spatiotemporal cuboids and employ 3D attention mechanisms to facilitate rich, joint modeling of spatial and temporal dependencies. This formulation demonstrates remarkable flexibility and scalability, accommodating variable resolutions and arbitrary video lengths with consistent performance. Nevertheless, the majority of current DiT-based approaches are conditioned exclusively on images and text, and do not incorporate audio as an explicit control signal—rendering them cannot be directly applied to audio-driven avatar generation.

\textbf{Audio-driven Avatar Generation.}
The avatar video generation task typically takes as input a reference portrait image of a person and an audio clip, and synthesizes a talking-head video of that synchronized with the provided speech. With the steady advancement in the capabilities of video foundation models, this field has experienced rapid progress. Early approaches were typically limited to animating human portrait images, commonly referred to talking-head generation, with representative methods including EMO \cite{tian2024emo}, VASA \cite{xu2024vasa}, loopy \cite{jiangloopy}. However, such methods are confined to facial animation and thus exhibit limited applicability in scenarios requiring full-body or more dynamic human representations. Subsequently, a line of works \cite{lin2025cyberhost,tian2025emo2,meng2025echomimicv3} have extended the input representation from talking-head portraits to upper-body (half-body) avatars. In these approaches, the audio conditioning signal governs not only lip synchronization but also hand and arm gestures, thereby enabling more expressive and holistic audio-driven animation. This advancement substantially enhances the behavioral richness and perceptual fidelity of avatar synthesis. Recently, a series of approaches \cite{cui2025high,fei2025skyreels,gan2025omniavatar,cui2025hallo3,wang2025fantasytalking,gao2025wan,jiang2025omnihuman} leveraging powerful text-to-video and image-to-video foundation models have enabled full-body avatar generation, unlocking compelling applications such as character interaction \cite{wang2025interacthuman} and multi-person dialogue \cite{kong2025let,wei2025mocha}. However, the motions of these generated avatars remain relatively simplistic and lack fine-grained controllability; in particular, they are hard to flexibly steered by text instructions to produce diverse movements or dynamic camera operations. As a result, these methods still fall short of the expressiveness, versatility, and cinematic quality.

\section{Method}
\label{sec:method}

\subsection{Overview}
\begin{figure}[tb]
  \centering 
  \includegraphics[width=\textwidth,trim=1.0cm 0.8cm 1.0cm 2.2cm,clip]{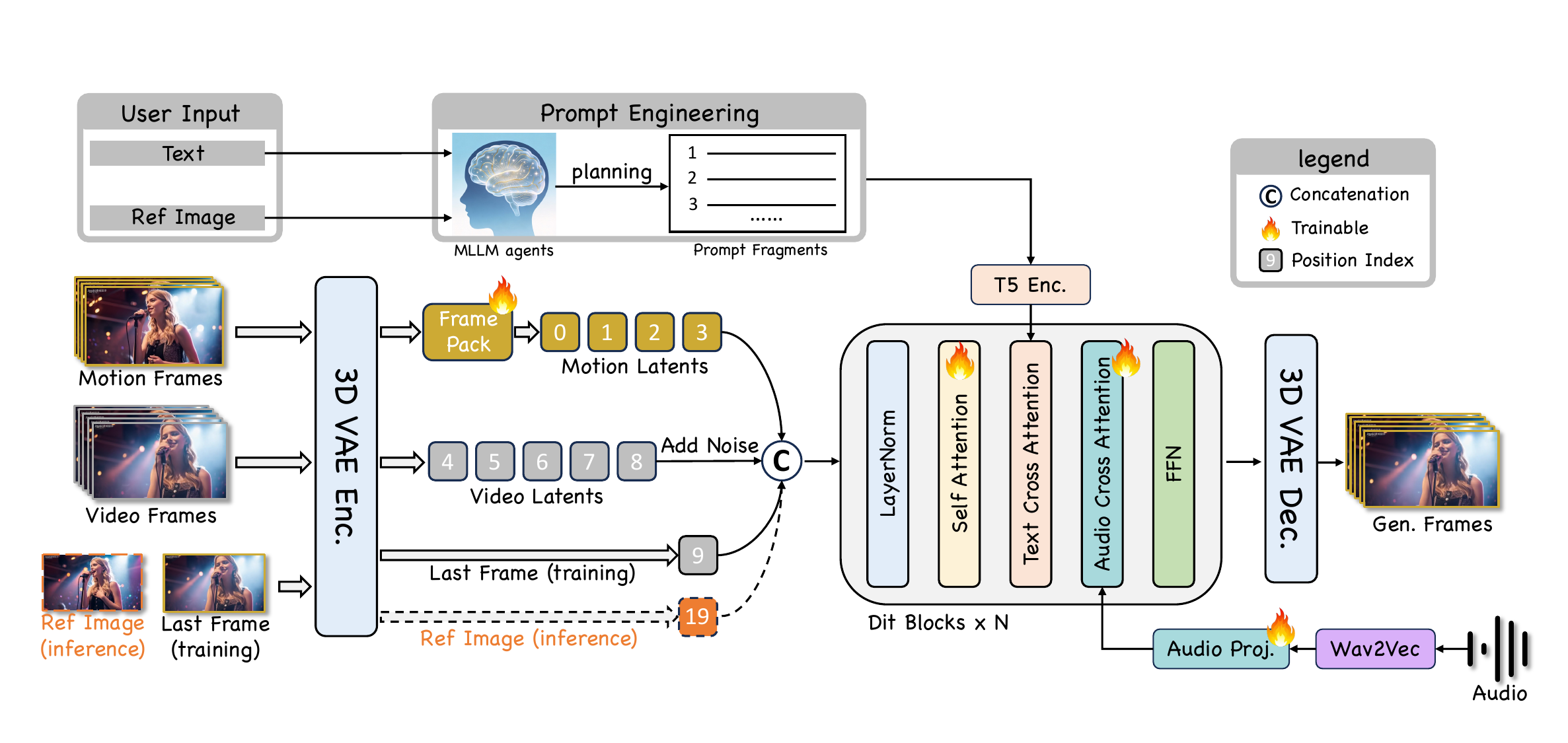}
  \caption{Overview of our JoyStreamer pretraining pipeline (Yellow-bordered frames denote noise-free frames while gray-bordered frames indicate noisy frames).}
  \label{fig:pipeline-01}
\end{figure}

Our goal is to generate avatar videos with faithful adherence to text instructions, long duration and precise lip-sync accuracy. To achieve this, we design the model architecture illustrated in Figure~\ref{fig:pipeline-01}. Specifically, we adopt the FramePack method \cite{zhang2025packing} to encode motion frames, ensuring seamless transitions between consecutive video segments. We adopt pseudo last frame strategy \cite{jiang2025omnihuman} to inject identity information from the reference image while avoiding error accumulation. Crucially, we propose two dedicated training strategies to ensure the avatar model's text controllability.

\subsection{Model Pretraining}
\label{sec_pretraining}
We first employ a 3D VAE \cite{wan2025wan} to map video clips into a compact latent space, and add noise to the video latents excluding the last frame. Subsequently, we concatenate the motion latents produced by the FramePack module, the noisy video latents (excluding the last frame), and the latent representation of the last frame along the temporal dimension, and feed this combined latent sequence into the foundation model, a DiT-based video diffusion model \cite{wan2025wan}. Then the model is trained using the flow matching objective \cite{liu2023flow}.

The overall model pretraining process consists of two stages. In the first stage, we train only the FramePack module and the self-attention layers of the foundation model, with the purpose of obtaining a base model capable of generating videos of unlimited duration. In the second stage, building on the base model obtained from the first stage, we insert audio cross-attention layers to capture the correlation between the visual and audio modalities. Specifically, a pretrained Wav2Vec2 \cite{wav2vec} is utilized to encode audio features. Then these audio features are injected into the base model through the newly introduced audio cross-attention layers, enabling precise audio-driven lip synchronization. Following these two training stages, the model obtains fundamental avatar functionalities and will serve as the audio-controllable teacher model in the subsequent DMD \cite{yin2024one} post-training phase.

\subsection{Model Post-training}
\label{sec_post-training}
The primary objective of this training phase is to enhance the text-controllability of the avatar video model. As a beneficial side effect, it also reduces the denoising steps, thereby improving computational efficiency during inference.

\subsubsection{Twin-Teacher Enhanced DMD Post-Training.}
\begin{figure}[tb]
  \centering 
  \includegraphics[width=\textwidth,trim=0.8cm 2.6cm 0.6cm 2.5cm,clip]{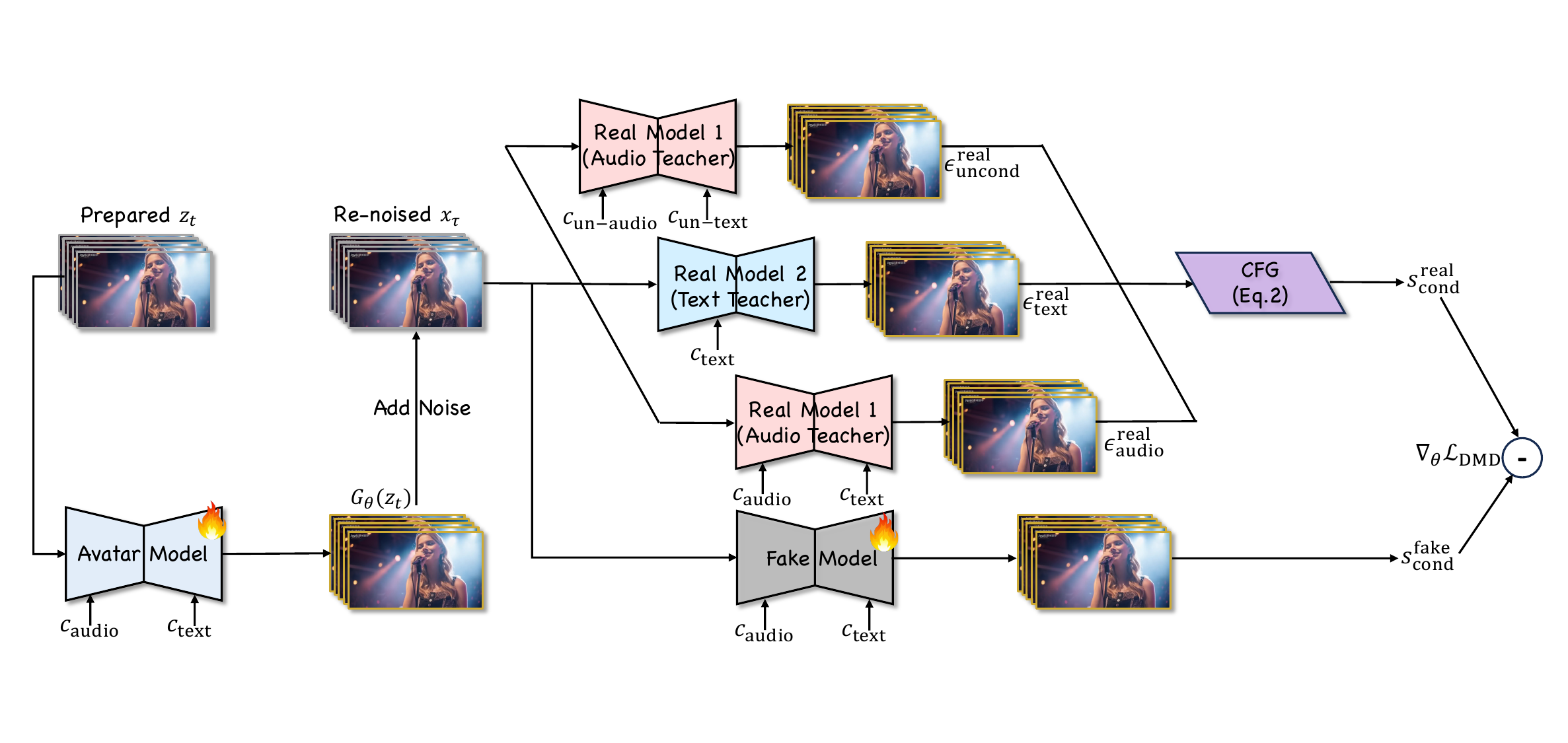}
  \caption{Pipeline of Twin-Teacher Enhanced DMD Post-Training.}
  \label{fig:pipeline-02}
\end{figure}

During the pretraining phase, we observe that the text-controllability particularly on motions of the avatar video model gradually degrades as training progresses. This degradation is likely due to the model overfitting to the relatively static style present in the training data, which consists overwhelmingly of videos featuring real humans engaged in speaking, lecturing, or singing. To address this issue, a straightforward solution is to curate a large-scale audio-visual dataset encompassing pronounced human motion and dynamic camera movements. However, such data is exceptionally hard to collect and is frequently filtered during standard data cleaning stages, as it often fails in video quality or consistency. To this end, we propose to leverage a pretrained video foundation model as another teacher and explicitly encourage the student avatar model to transfer the diverse style text-controllability from video foundation model. Since the DMD framework inherently incorporates a teacher model, i.e., real score function, as a core component, we implement this idea within the DMD framework. In formal, the DMD objective produces the gradient with respect to the avatar generator parameters $\theta$ as:
\begin{equation}
\nabla_\theta \mathcal{L}_{\text{DMD}} = \mathbb{E}_{z_t, \tau, \mathbf{x}_\tau} \left[ -\left( s^{\text{real}}_{\text{cond}}(\mathbf{x}_\tau) - s^{\text{fake}}_{\text{cond}}(\mathbf{x}_\tau) \right) \frac{\partial G_\theta(z_t)}{\partial \theta} \right],
\label{eq1}
\end{equation}
where $G_\theta$ denotes avatar generator, $z_t$ denotes the input of $G_\theta$ at noise level $t$, $\mathbf{x}_\tau$ denotes the result obtained by renoising to the denoised output by $G_\theta$ at noise level $\tau$. In practice, inspired by \cite{liu2025decoupled}, we correspondingly define $s^{\text{real}}_{\text{cond}}(\mathbf{x}_\tau)$ in Eq.~\ref{eq1} as follows:
\begin{equation}
\begin{split}
s^{\text{real}}_{\text{cond}}(\mathbf{x}_\tau) &= \epsilon_{\text{audio}}(\mathbf{x}_\tau, \tau, c_{\text{un-text}}, c_{\text{un-audio}})
 \\ 
& +\alpha_a \cdot \left(\epsilon_{\text{audio}}(\mathbf{x}_\tau, \tau, c_{\text{text}}, c_{\text{audio}}) - \epsilon_{\text{audio}}(\mathbf{x}_\tau, \tau, c_{\text{text}}, c_{\text{un-audio}}) \right)  \\
& +\alpha_t \cdot \left(\epsilon_{\text{audio}}(\mathbf{x}_\tau, \tau, c_{\text{text}}, c_{\text{un-audio}}) - \epsilon_{\text{audio}}(\mathbf{x}_\tau, \tau, c_{\text{un-text}}, c_{\text{un-audio}}) \right).
\label{eq1-regular-dmd}
\end{split}
\end{equation}
$\epsilon_{\text{audio}}$ is the base model obtained in pretraining phase (see Sec.~\ref{sec_pretraining}), $c_{\text{text}}$ and $c_{\text{audio}}$ are text conditions and audio conditions, respectively. $c_{\text{un-text}}$ represents the negative prompt condition. $c_{\text{un-audio}}$ typically denotes a zero embedding (null audio condition). $\alpha_t$ and $\alpha_a$ are the text CFG guidance scale and audio CFG guidance scale. Given that $\epsilon_{\text{audio}}$ is inherently limited in providing accurate supervision for text-conditioned generation, we incorporate the pretrained Wan2.2-I2V-14B model \cite{wan2025wan} as another teacher specifically designed to deliver expert-level supervision on text alignment. Consequently, we redefine $s^{\text{real}}_{\text{cond}}(\mathbf{x}_\tau)$ in Eq.~\ref{eq1} as follows:
\begin{equation}
\begin{split}
s^{\text{real}}_{\text{cond}}(\mathbf{x}_\tau) &= \chighlight{red!20}{\epsilon_{\text{audio}}}(\mathbf{x}_\tau, \tau, c_{\text{un-text}}, c_{\text{un-audio}})
 \\ 
& +\alpha_a \cdot \underbrace{\left(\chighlight{red!20}{\epsilon_{\text{audio}}}(\mathbf{x}_\tau, \tau, c_{\text{text}}, c_{\text{audio}}) - \chighlight{cyan!20}{\epsilon_{\text{text}}}(\mathbf{x}_\tau, \tau, c_{\text{text}}) \right)}_{\color{BrickRed}{\text{audio condition engine}}}  \\
& +\alpha_t \cdot \underbrace{\left(\chighlight{cyan!20}{\epsilon_{\text{text}}}(\mathbf{x}_\tau, \tau, c_{\text{text}}) - \chighlight{red!20}{\epsilon_{\text{audio}}}(\mathbf{x}_\tau, \tau, c_{\text{un-text}}, c_{\text{un-audio}}) \right)}_{\color{RoyalBlue}{\text{text condition engine}}}.
\label{eq2}
\end{split}
\end{equation}
$\chighlight{red!20}{\epsilon_{\text{audio}}}$ is the base model obtained in pretraining phase (see Sec.~\ref{sec_pretraining}), termed the audio teacher. $\chighlight{cyan!20}{\epsilon_{\text{text}}}$ is the newly introduced teacher, i.e., the pretrained Wan2.2-I2V-14B model, termed the text teacher. As vividly illustrated in Figure~\ref{fig:pipeline-02}, substituting Eq.~\ref{eq2} into Eq.~\ref{eq1}, and subsequently optimizing with DMD loss, the avatar model $G_\theta$ is expected to maximally transfer the faithful text-controllability from the video foundation model.

\subsubsection{Dynamic CFG Strategy.}
On the other hand, to effectively mitigate the modality conflict between audio and text conditions as much as possible, we draw inspiration from Alignhuman \cite{liang2025alignhuman} and investigate the impact of different modalities by examining their influence across distinct denoising timestep intervals. Alignhuman \cite{liang2025alignhuman} reveals that the global motion structure has been established in the high-noise (early) denoising steps. To this end, during the twin-teacher DMD post-training process, we dynamically adjust the text and audio CFG scales in Eq.~\ref{eq2}. Specifically, we set the audio CFG scale $\alpha_a$ to 0.1, when the denoising timestep exceeds $t_{\text{spot}}$ (typically set to 950), to prioritize the text condition. Correspondingly, we increase the audio CFG scale $\alpha_a$ to 4.0 to ensure the lip-sync accuracy when the denoising timestep is below $t_{\text{spot}}$. Such design temporally decouples the dominant periods of influence for text and audio conditioning and is expected to reduce inter-modal interference while preserving the strengths of both conditions.

In practice, we observed that models trained solely with Eq.~\ref{eq2} exhibit significantly enhanced responsiveness to text instructions but suffer from reduced temporal stability in the generated videos. To address this trade-off, after completing the Twin-Teacher Enhanced DMD post-training phase, we perform an additional round of vanilla DMD training, based on Eq.~\ref{eq1-regular-dmd}, to preserve visual-temporal coherence and overall video stability.

\subsection{Training-Free Audio-driven Multi-Person Dialogues}
This subsection investigates the applicability of our avatar model to multi-person conversational application. While some existing methods\cite{kong2025let} can address the problem of generating multi-person videos driven by multiple audio streams, they require extensive multi-person video training data, leading to high training costs and limited scalability to larger group sizes. Additionally, these methods cannot handle multi-round conversations effectively. In this section, we develop a training-free multi-person animation algorithm, which can be rapidly adapted to any audio-driven video generation base model. Moreover, the proposed algorithm supports conversations involving more than three participants and enables seamless handling of multi-round dialogues.

\begin{figure}[tb]
  \centering 
  \includegraphics[width=\textwidth]{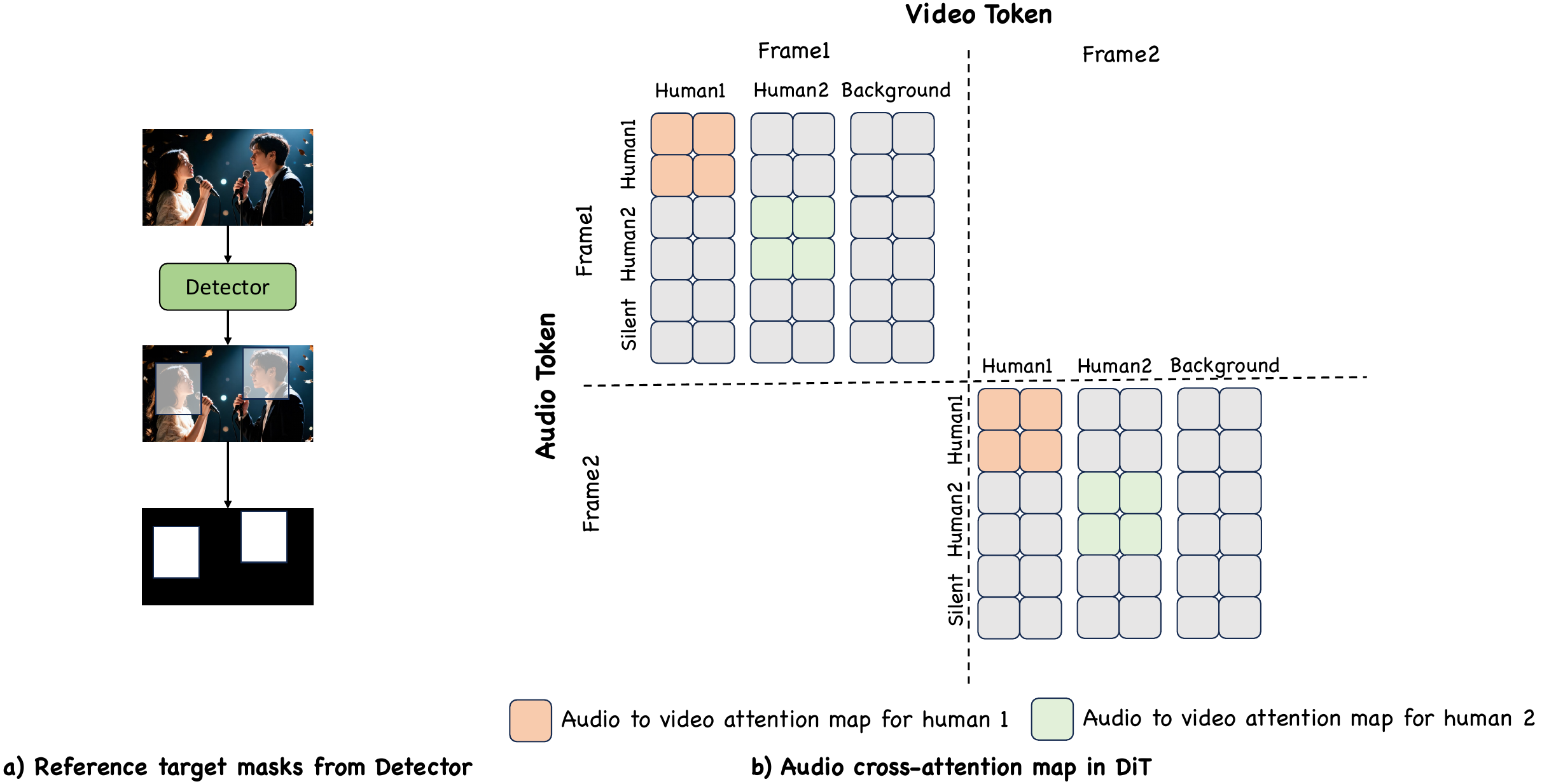}
  \caption{Pipeline of Multi-Person audio across-attention.}
  \label{fig:multi-talk}
\end{figure}

As illustrated in Figure~\ref{fig:multi-talk}(a), the detector module is used to identify the main objects in the image. Given a binary mask map, set the detected regions to 1 and the rest to 0. In the audio cross-attention mechanism, queries are derived from the video latent, whereas keys and values originate from the multi-stream audio embeddings. As depicted in Figure~\ref{fig:multi-talk}(b), taking the first human as an example, the audio token of human1 will only perform audio cross-attention calculations with the reference target masks for human1. This method is applied for each person in the same manner. For regions where the mask is set to zero, a silent audio is utilized to perform cross-attention with the corresponding regions during the audio-based computation. Using the aforementioned operation, we accomplish the creation of multi-character audio-driven videos without requiring any training phase.

\section{Experiments}
\subsection{Implementation Details}
Our avatar model is built upon the Wan2.2-T2V-14B \cite{wan2025wan} foundation model. We set the length of motion frames to 9 frames and that of video frames to 80 frames. During the pretraining phase, we separately train the avatar models based on high-expert foundation model and low-expert foundation model derived from the Wan2.2 model. We employ the AdamW optimizer with a learning rate of 2e-5, a global batch size of 64. In the post-training phase, based on the model obtained by pretraining phase, we fine-tune the audio projection layer, audio cross-attention layers, and self-attention layers using the LoRA strategy, with a learning rate of 1e-5. During training, we utilize our self-collected dataset comprising approximately 4,000 hours of video data, encompassing speaking, speech delivery, and singing by both humans and cartoon characters.

During inference, as illustrated in Figure~\ref{fig:pipeline-01}, we treat the input reference image as the pseudo last frame and adjust its positional index by adding 10 to the original index. This adjustment prevents the model from strictly reverting to the last frame and enables a richer motion dynamics. Upon inference completion, the latent representations associated with the motion frames and the pseudo last frame are discarded. As a result, each inference pass produces 76 available video frames. To avoid repetitive responses to the same input prompt across generated video segments, we optimize the PE (Prompt Engineering) module based on Qwen3-VL-32B \cite{bai2025qwen3vltechnicalreport}. The optimized PE module is designed to adaptively produce a variable number of text prompt fragments according to the duration of the user’s audio input. Specifically, for every 3.24 seconds of input audio, the PE model generates an additional prompt fragment, up to a maximum of five fragments. By analyzing the temporal structure and semantics of the user’s instructions, the model distributes time-ordered prompts across different segments appropriately, minimizing repetitive actions.

\subsection{Qualitative Comparison}
We adhere to the human preference–based subjective evaluation protocol to perform a comprehensive assessment of our model. We construct a test set comprising 96 carefully curated cases, each consisting of a reference image, an audio clip, and a text instruction. Some of the image and audio assets used are courtesy of \cite{zhang2025soul}. The text instructions encompass a diverse and challenging range of semantic content—including facial expressions, body motions, human-object interactions, camera movements, and background transitions—designed to thoroughly evaluate the model’s fidelity in following complex text prompts. Additionally, the test set includes numerous cases with audio durations exceeding 30 seconds, enabling a robust assessment of temporal stability and visual consistency over extended generation horizons. For each test case, we ask human evaluators to perform pairwise comparisons between our method and each competitive method, selecting one of three possible outcomes: (1) our method is better, (2) the competitive method is better, or (3) both methods are perceptually equivalent. We then compute the evaluation score as (G + S) / (B + S), where G denotes the number of good for our method, B the number of bad for our method, and S the number of the same. A higher score indicates superior performance. To ensure a comprehensive assessment, evaluations are conducted across the following five dimensions:
\begin{itemize}
    \item \textbf{Overall Quality}. Assesses whether the generated video appears natural and realistic, adhering to physical plausibility and visual coherence.
    \item \textbf{Text Alignment}. Evaluates the model’s ability to faithfully execute the specified instructions in the text prompt, including facial expressions, body motions, camera movement trajectories.
    \item \textbf{Lip Synchronization}. Assesses whether the lip movements accurately reflect the phonetic content and timing of the input speech.
    \item \textbf{Identity Consistency}. Evaluates whether the identity of the generated character remains faithful to the input reference image throughout the video, with particular attention to long-duration sequences where identity drift may occur.
    \item \textbf{Visual Quality}. Assesses the perceptual quality of the generated video, focusing on image sharpness, textural detail, and the absence of artifacts or synthetic-looking characteristics commonly associated with AI-generated content (i.e., "AIGC look").
\end{itemize}
\begin{figure}[tb]
    \centering
    \begin{minipage}{0.48\textwidth}
        \centering
        \includegraphics[width=\linewidth]{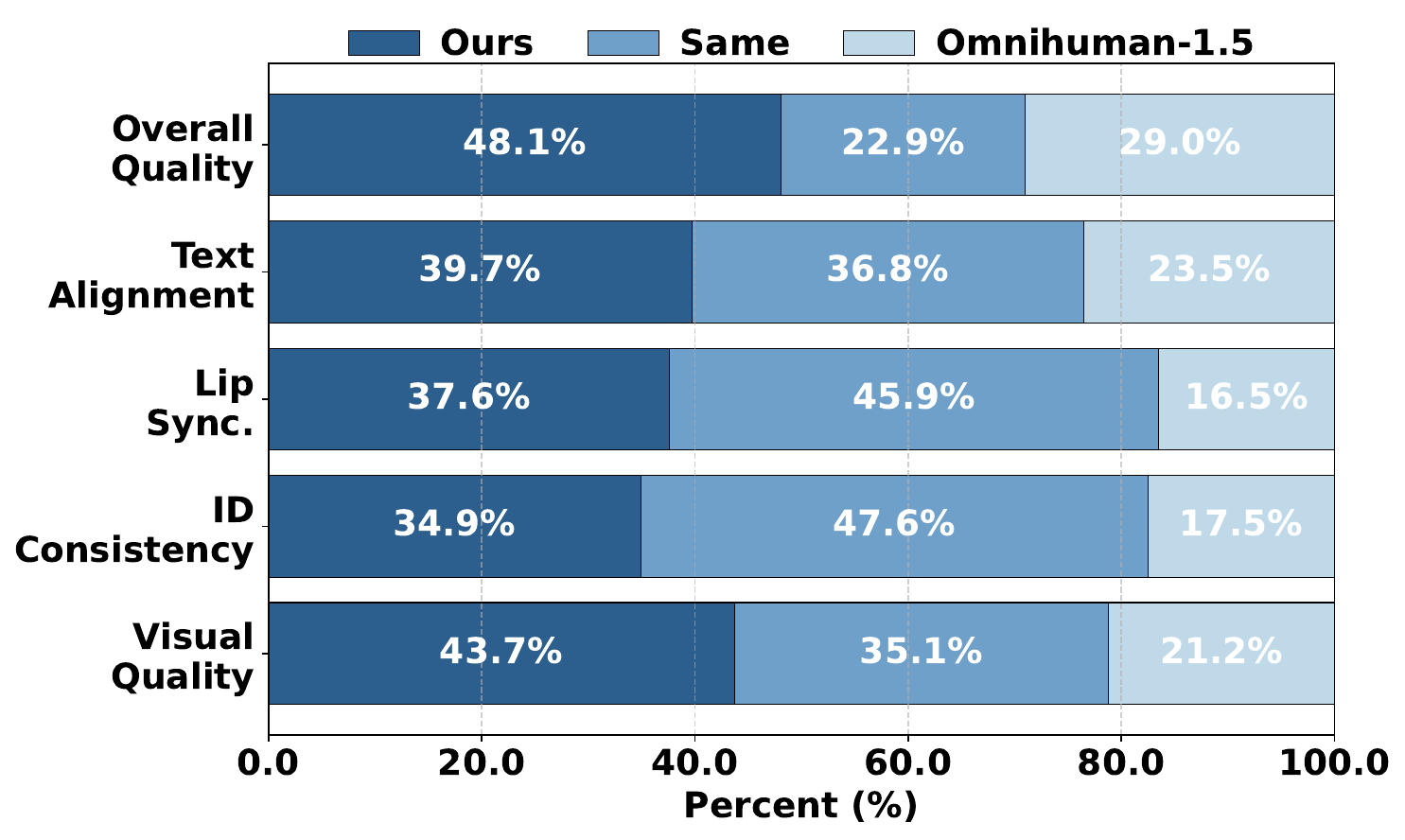}
    \end{minipage}
    \begin{minipage}{0.48\textwidth}
        \centering
        \includegraphics[width=\linewidth]{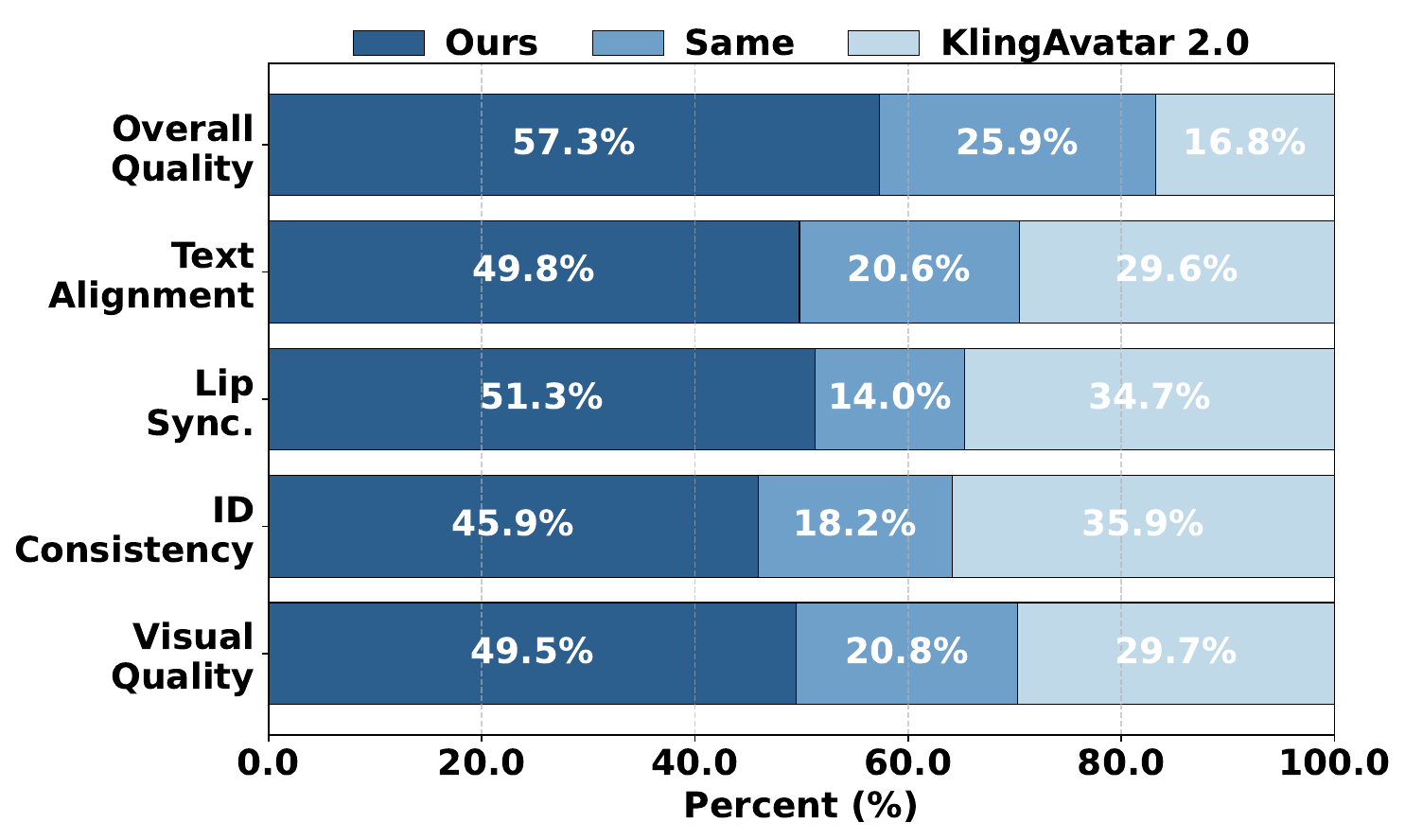}
    \end{minipage}
    \caption{GSB comparison results between our model and other competitors across various evaluation dimensions.}
    \label{fig:comparison-with-sota-experiments}
    \vspace{-0.3cm}
\end{figure}

\begin{figure}[tb]
    \centering 
    \includegraphics[width=0.98\textwidth,trim=3.0cm 7.0cm 1.0cm 2.0cm,clip]{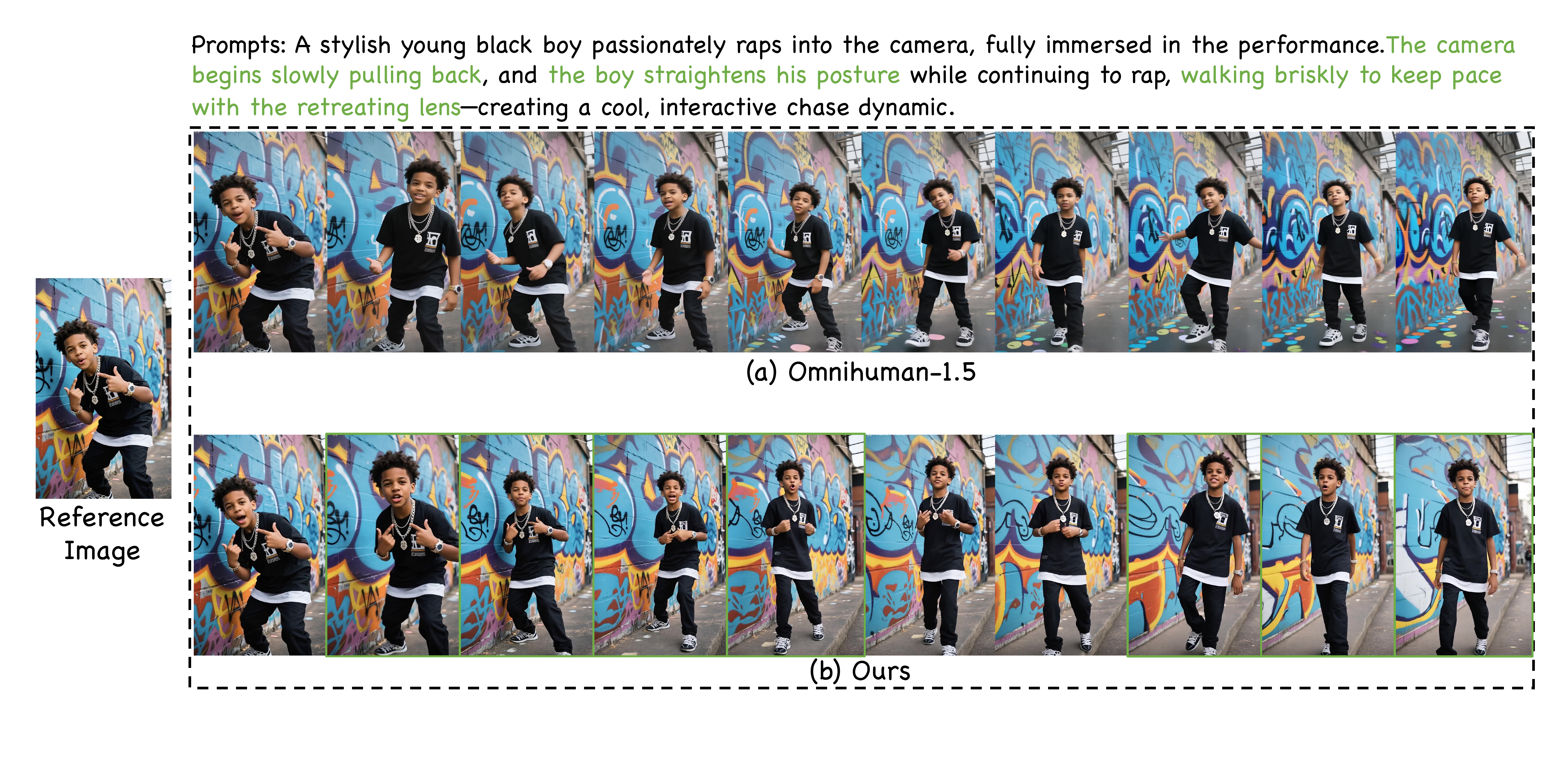} \\
    \includegraphics[width=0.98\textwidth,trim=3.0cm 6.0cm 0.0cm 7.0cm,clip]{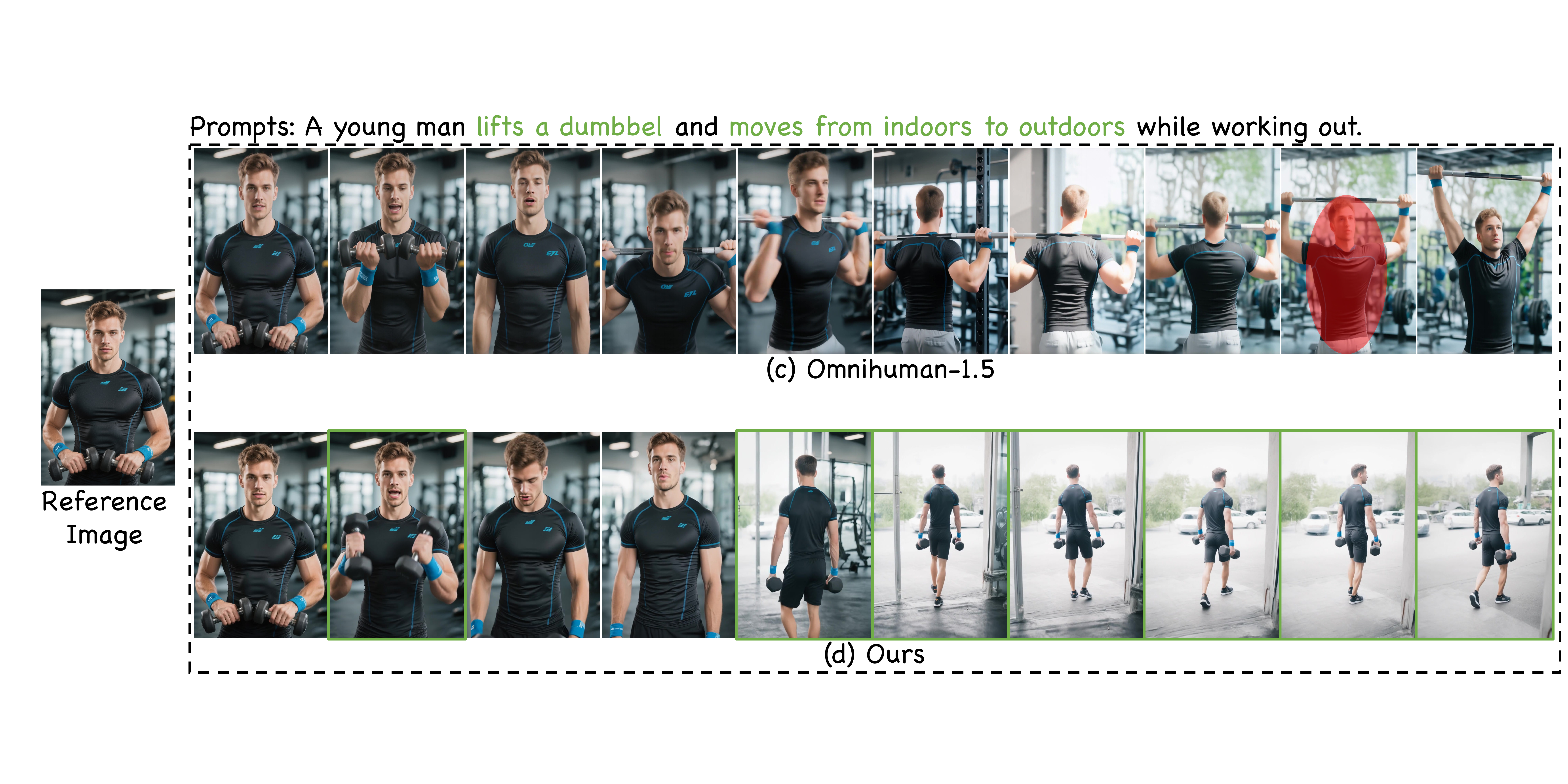}
    \caption{Some issues of omnihuman-1.5: AIGC "look" in (a) and abrupt jumps in (c) (Key actions in the prompts and correct response in generated videos are marked in green).}
    \label{fig:our-compare-with-omni}
\end{figure}
\begin{figure}[tb]
    \centering 
    \includegraphics[width=0.98\textwidth,trim=2.0cm 7.0cm 1.0cm 7.0cm,clip]{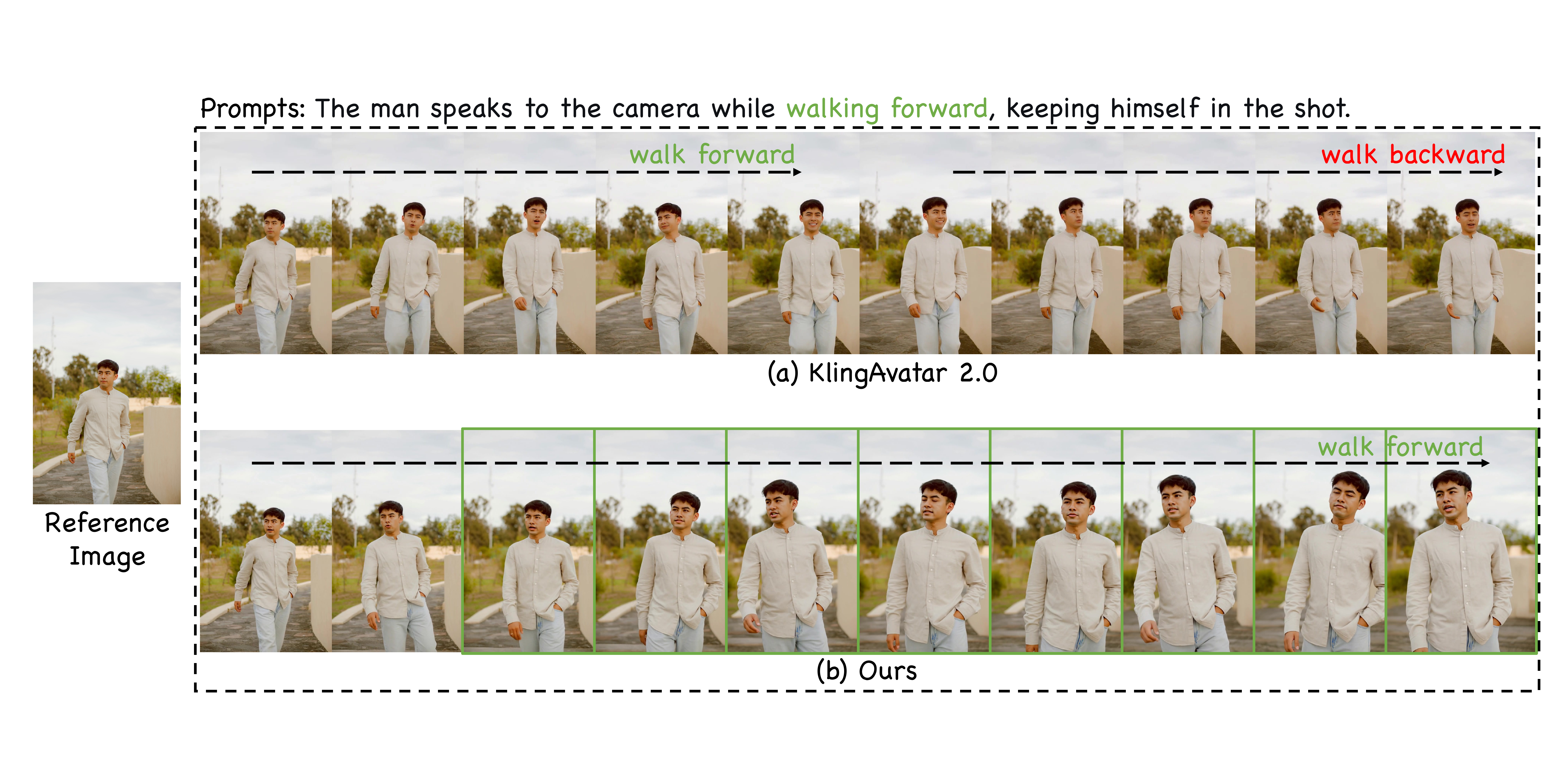} \\
    \includegraphics[width=0.98\textwidth,trim=2.0cm 7.0cm 1.0cm 9.5cm,clip]{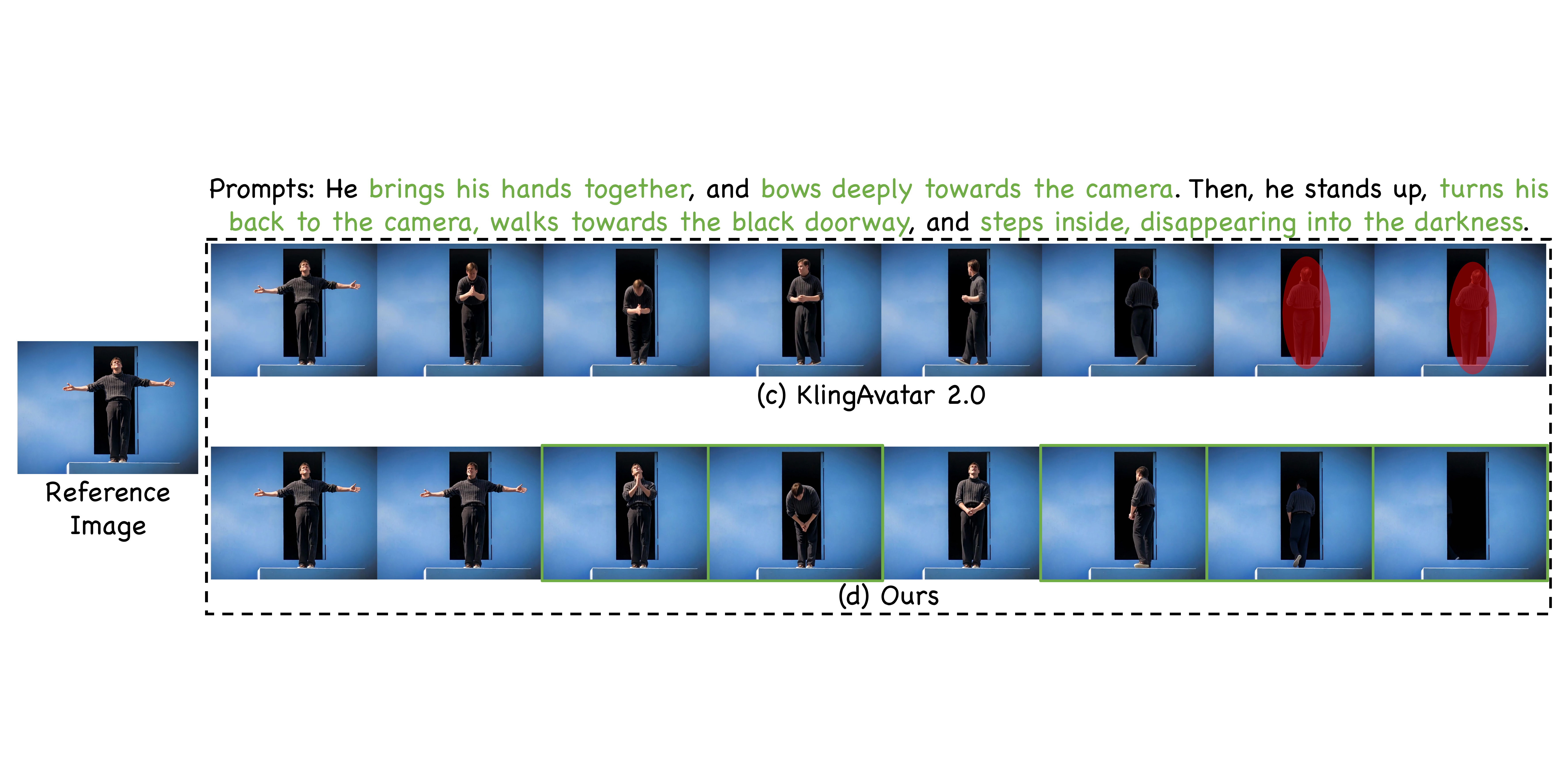}
    \caption{Some issues of KlingAvatar 2.0: abruptly reversing the direction in (a) and (c) (Key actions in the prompts and correct response in generated videos are marked in green while problematic regions in the generated video are highlighted in red).}
    \label{fig:our-compare-with-kling}
\end{figure}

\begin{table}[t]
\centering
\caption{Quantitative comparisons with several representative avatar models.}
\label{tab:with-sota}
\begin{tabular}{lcccccc}
\toprule
\multicolumn{1}{c}{Methods} & \multicolumn{6}{c}{Metrics} \\
\cmidrule(lr){2-7}
& Sync-D$\downarrow$ & Sync-C$\uparrow$ & IDC$\uparrow$ & Q-Score$\uparrow$ & HKC$\uparrow$ & HKV$\uparrow$\\
\midrule
Omnihuman-1.5\cite{jiang2025omnihuman} & 9.31 & 4.31 & 0.71 & \textbf{0.92} & 0.87 & 48.41\\
KlingAvatar 2.0\cite{team2025klingavatar} & 8.97 & 4.39 & 0.77 & 0.92 & 0.87 & 47.65 \\
HeyGen\cite{heygen} & 9.09 & 4.86 & 0.79 & 0.92 & 0.86 & \textbf{58.76} \\
Wan-S2V\cite{gao2025wan} & 9.34 & 3.96 & 0.71 & 0.91 & 0.87 & 35.24 \\
Ours & \textbf{8.09} & \textbf{5.57} & \textbf{0.79} & 0.91 & \textbf{0.87} & 32.15\\
\bottomrule
\end{tabular}
\end{table}
We compare our method against two representative closed-source models: Omnihuman-1.5 \cite{jiang2025omnihuman} and KlingAvatar 2.0 \cite{team2025klingavatar}. The GSB (Good–Same–Bad) evaluation results are visualized in Figure~\ref{fig:comparison-with-sota-experiments}. As shown, our approach consistently outperforms both competitors across all five evaluation dimensions. Specifically, we observe that the KlingAvatar 2.0 model frequently exhibits violations of physical plausibility—particularly in scenes involving human motion. For instance, in cases where a character is walking forward, the motion often abruptly reverses direction after a few seconds (See Figure~\ref{fig:our-compare-with-kling}(a)(c)). We speculate that this artifact stems from KlingAvatar 2.0’s underlying generation paradigm, which appears to resemble blueprint keyframes (e.g., generating video segments based only on start and end frames). Such a design lacks explicit modeling of motion frames, leading to temporally fragmented outputs where consecutive chunks are generated independently, thereby compromising physical coherence. Moreover, KlingAvatar 2.0 exhibits limited camera movements. As shown in the results shown in Figure~\ref{fig:comparison-with-sota}, Omnihuman-1.5 outperforms KlingAvatar 2.0; however, we find several notable limitations of this model. For instance, the generated videos display perceptible AI-generated artifacts—commonly referred to as an “AIGC look” (See Figure~\ref{fig:our-compare-with-omni}(a)). Moreover, Omnihuman-1.5 frequently introduces camera movements not specified in the input text prompts, often accompanied by abrupt jumps, or visual artifacts (See Figure~\ref{fig:our-compare-with-omni}(c)). Finally, as illustrated in Figure~\ref{fig:complex-text-aligment}, both Omnihuman-1.5 and KlingAvatar 2.0 exhibit limited fidelity in following complex text instructions—particularly in scenarios involving human object interactions—indicating substantial room for improvement in the text alignment. Specifically, in Figure~\ref{fig:complex-text-aligment}(a), Omnihuman-1.5 fails to generate the action of the person eating chocolate, while KlingAvatar 2.0 does not render the chocolate object at all. In Figure~\ref{fig:complex-text-aligment}(b), Omnihuman-1.5 does not depict the person climbing the ladder as instructed, and KlingAvatar 2.0 generates a motion trajectory in which the person walks in an incorrect direction, away from the direction of the ladder. These examples highlight significant shortcomings in both models’ ability to accurately execute complex, semantics-rich instructions.

\subsection{Quantitative Comparison}
In addition to evaluating the model’s ability to follow complex text prompts, we also construct a custom benchmark specifically designed to assess the fundamental capabilities of ours avatar model, such as speaking and delivering speeches, in relatively static scenarios. This benchmark comprises 90 carefully curated cases, encompassing speaking, singing, and speech delivery by a diverse range of characters, including humans, animals, cartoons, and dolls. In this benchmark, we primarily focus on key metrics, such as lip-sync accuracy and identity consistency (IDC), and give a trivial test prompt. Specifically, for comparative methods, we add one closed-source model, HeyGen\cite{heygen}, and one representative open-source model, Wan-S2V\cite{gao2025wan}. In terms of evaluation metrics, we employ SyncNet’s \cite{syncnet} Sync-C score and Sync-D (lip distance) to quantitatively measure audio–visual synchronization fidelity. Identity consistency of the human between the generated video and the reference image is evaluated via cosine similarity computed on DINOv2 \cite{oquab2023dinov2} features. Additionally, we adopt HKC (Hand Keypoint Confidence) and HKV (Hand Keypoint Variance), proposed in \cite{lin2025cyberhost}, to assess hand confidence and hand motion richness, respectively. Quantitative comparison results are summarized in Table~\ref{tab:with-sota}.

As shown in Table~\ref{tab:with-sota}, our model significantly outperforms competitive methods on key metrics such as lip-sync accuracy and identity consistency. However, it exhibits slightly lower HKV compared to Omnihuman-1.5 and KlingAvatar 2.0. This is because, when test cases with trivial text prompts, we adjust the positional index of pseudo last frame by adding 3 to the original index. This adjustment was made to minimize extraneous human motion with trivial text prompts.

\subsection{Ablation Study}
\begin{figure}[tb]
  \centering 
  \includegraphics[width=\textwidth,trim=0.0cm 0cm 0.0cm 0.0cm,clip]{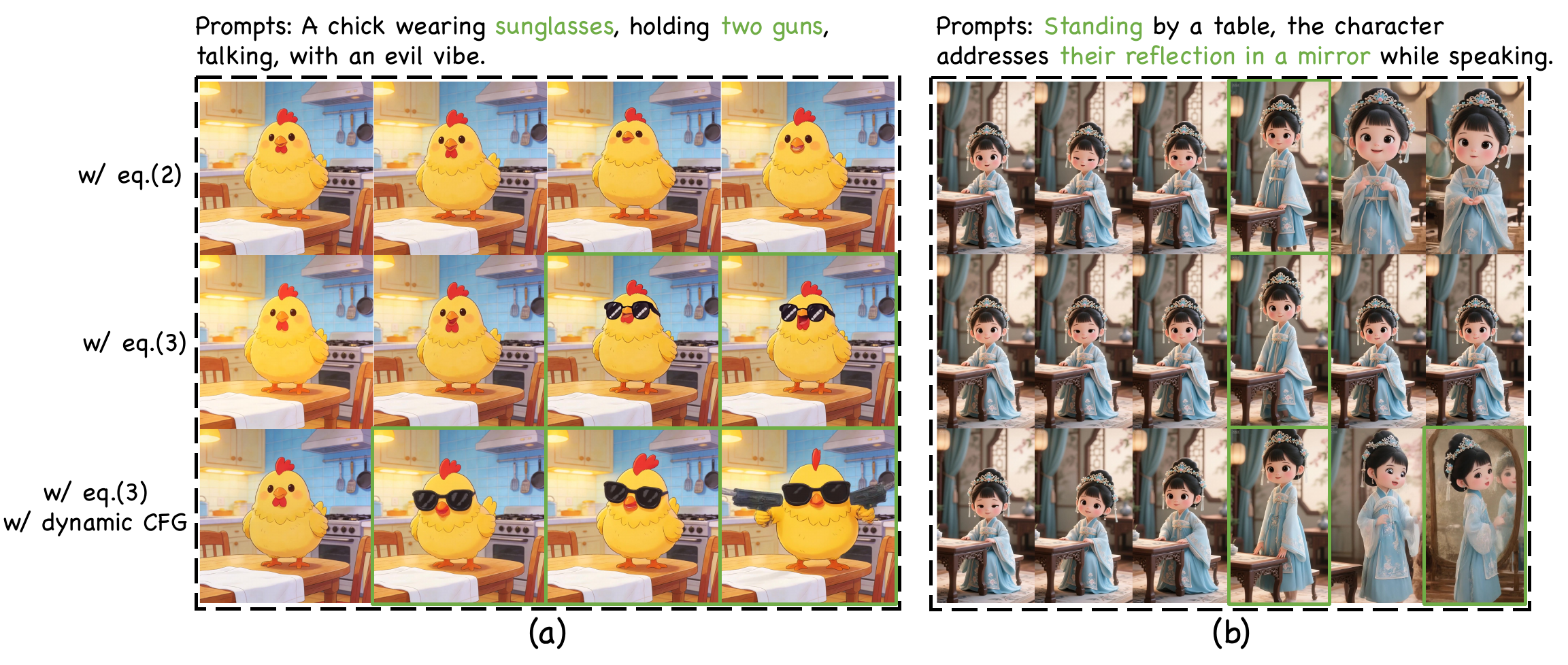}
  \caption{Ablation study of the twin-teacher enhanced DMD post-training and dynamic CFG strategy in training process.}
  \label{fig:ablation}
\end{figure}

We primarily investigate the impact of our proposed training innovations on the text-alignment capability of video avatar models. As illustrated in Figure~\ref{fig:ablation}, both of our proposed designs improve text alignment over the baseline model, with particularly in responsiveness to novel object. Moreover, in practice, we find that after twin-teacher–enhanced DMD training, the model exhibits significantly enhanced motion dynamism compared to the baseline. We believe that the limited lexical and semantic diversity in current talking-style datasets, whose text annotations are often rigid, constrains the full potential of the twin-teacher training mechanism. With richer, more comprehensive text annotations, this training paradigm holds substantial promise for further advancing text-to-motion alignment fidelity.


\section{Conclusion}
In this work, we introduce JoyStreamer model, which enables comprehensive text-driven control over avatar articulation, cinematic camera movements, and multi-character conversations, while supporting the generation of long duration. Our key innovation lies in enhancing the text-driven controllability of video avatar models from a novel perspective. Specifically, we firstly introduce twin-teacher enhanced DMD post-training strategy to maximally transfer inherent text control capabilities from the video foundation model. Secondly, we propose a dynamic classifier-free guidance (CFG) mechanism that adaptively modulates conditioning signals across distinct denoising timesteps to mitigate conflict between heterogeneous control modalities. Experimental results demonstrate that our model not only achieves precise audio–visual synchronization but also enables avatar to perform complex human tasks and interactive behaviors, thereby significantly advancing overall expressiveness toward lifelike realism.


%
%
\bibliographystyle{splncs04}
\bibliography{main}
\end{document}